\documentclass[lettersize,journal]{IEEEtran}

\usepackage{graphicx}
\usepackage{epstopdf}
\usepackage{multirow,diagbox}
\usepackage{dirtytalk}
\usepackage{caption}
\usepackage{subcaption}
\usepackage{amsmath,nccmath,bm,mathtools}

\IEEEoverridecommandlockouts                              

\begin{document}

\title{\LARGE \bf
Steady-Hand Eye Robot 3.0: Optimization and Benchtop Evaluation for Subretinal Injection
}

\author{Alireza Alamdar$^{*1}$, David E. Usevitch$^{*1}$, \IEEEmembership{Student Member,~IEEE, } Jiahao Wu $^{1}$,  Russell H. Taylor$^{1}$, \IEEEmembership{Life Fellow,~IEEE, }  Peter Gehlbach$^{2}$, and Iulian Iordachita$^{1}$, \IEEEmembership{Senior Member,~IEEE, }
\thanks{* Alireza Alamdar and David E. Usevitch acted as co-first authors.}
\thanks{$^{1}$Laboratory for Computational Sensing and Robotics (LCSR), Johns Hopkins University, Baltimore, MD USA, Corresponding authors: David Usevitch  {\tt\small usevitch@jhu.edu} and Iulian Iordachita {\tt\small iordachita@jhu.edu}}%
\thanks{$^{2}$Wilmer Eye Institute, Johns Hopkins University, Baltimore, MD 21287, USA}%
\thanks{This research was supported by the U.S. National Institutes of Health under grant number 1R01EB025883-01A1 and Johns Hopkins University internal funds.}}

\maketitle

\begin{abstract}
Subretinal injection methods and other procedures for treating retinal conditions and diseases (many considered incurable) have been limited in scope due to limited human motor control. This study demonstrates the next generation, cooperatively controlled Steady-Hand Eye Robot (SHER 3.0), a precise and intuitive-to-use robotic platform achieving clinical standards for targeting accuracy and resolution for subretinal injections. The system design and basic kinematics are reported and a deflection model for the incorporated delta stage and validation experiments are presented. This model optimizes the delta stage parameters, maximizing the global conditioning index and minimizing torsional compliance. Five tests measuring accuracy, repeatability, and deflection show the optimized stage design achieves a tip accuracy of $\boldsymbol{<30}$\,$\boldsymbol{\mu}$m, tip repeatability of $\boldsymbol{9.3}$\,$\boldsymbol{\mu}$m and 0.02$\boldsymbol{^\circ}$, and deflections between 20-350\,$\boldsymbol{\mu}$m/N. Future work will use updated control models to refine tip positioning outcomes and will be tested on \emph{in vivo} animal models.
\end{abstract}


\section{Introduction}

Vitreoretinal surgery is one of the most technically demanding types of eye surgery \cite{channa2017robotic} and treats retinal and posterior eye segment diseases. Following the trend in microsurgery \cite{mattos2016microsurgery} and enhanced by advanced imaging, robotic assistance has the potential to fundamentally change the field of intraocular surgery through tremor reduction, force scaling, and other precise teleoperative and cooperative control capabilities. Though still in its infancy, robotic retinal surgery has been cautiously introduced into the operating room over the last decade and has been successfully validated in a limited number of clinical trials \cite{de2018robotic, roizenblatt2018robot, vander2020robotic}. Nonetheless, owing to its demonstrated capabilities, robotic intraocular microsurgery has the potential to enable unprecedented surgical enhancements for surgeons and safer surgical care for patients. 

Consider the case of subretinal injection for treating retinal degenerative diseases such as age-related macular degeneration (AMD). AMD is currently the leading cause of blindness worldwide \cite{lim2012age, ratnapriya2013age} and is projected to affect 288 million people worldwide by 2040 \cite{WONG2014e106}. In AMD, the central vision progressively worsens as parts of the central retina degenerate leading to moderate or severe vision loss. There is currently no accepted cure, though several commercially available drugs are commonly used in treatment. Direct care costs to healthcare systems due to AMD amount to \$255 billion annually \cite{MacularDegenerationResearch}. Recent advances in regenerative medicine highlight potential clinical applications of gene therapy and stem cell therapy approaches to heal diseased RPE or photo-receptor cell layers \cite{schwartz2012embryonic, schwartz2015human, song2016looking, cai2006bruch, jager2008age} mainly via injection methods to deliver drugs to the subretinal space. 

\begin{figure}[!t]
    \centering
    \vspace{0.2cm}
    \includegraphics[width=\linewidth]{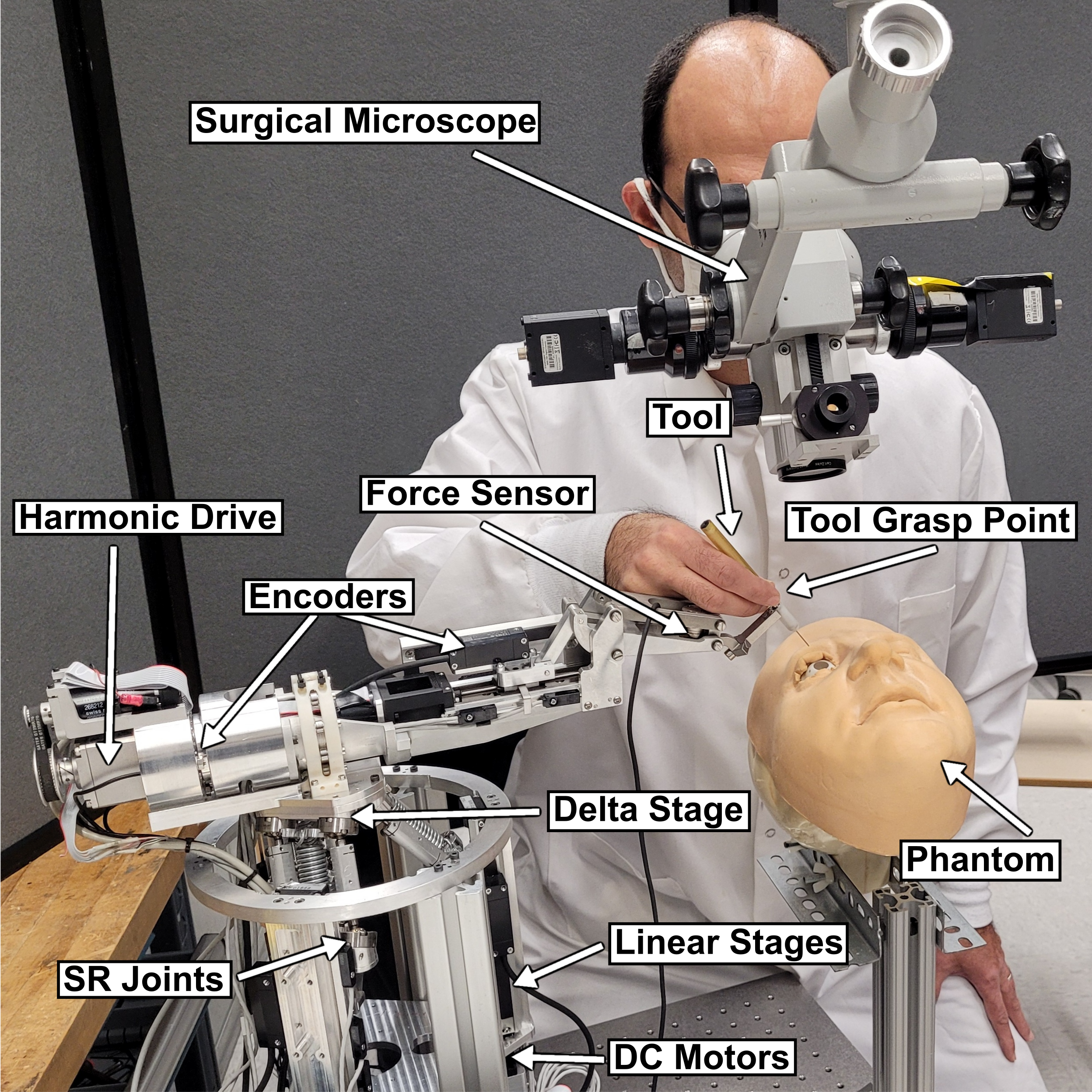}
    \caption{The cooperatively controlled Steady-Hand Eye Robot (SHER) 3.0 is guided by a user around a phantom eye model. This figure shows the robot setup with a surgical microscope, allowing ease of use in the operating room (OR). The SHER 3.0 can be easily moved for ergonomic use by a left- or right-handed surgeon. Its compact size makes it favorable for use in space-expensive ORs.}
    \label{fig: whole prototype}
\end{figure}

Despite advances in regenerative medicine, the translation of therapeutic delivery methods into an accepted standard of care for clinical use has been limited. This is partially due to the lack of effective ways to deliver these therapeutics quickly, safely, and efficiently to the fragile anatomy of the sub-retinal space.  Delivery risks include irreversible damage to the retina and ineffective delivery of the cell therapy \cite{ehlers2016intraoperative, wert2012subretinal, kang2003new, peng2017subretinal, hauswirth2008treatment}, among others. The appearance or avoidance of severe adverse effects from subretinal injection is influenced by delivery technique, delivery medium, and surgical precision and expertise. Poor outcomes reported have included adverse clinician physiological hand tremor \cite{novelli2017new}. The ability to make subretinal delivery more steady, consistent, and accurate could be significantly enhanced via robotic delivery devices that improve microsurgery in many other instances. Additionally, other vitreoretinal surgery applications (i.e., retinal vein cannulation) would become feasible and greatly benefit from using improved robotic surgery systems.

\begin{figure*}[!t] 
    \centering
    \vspace{0.2cm}
    \includegraphics[width=\textwidth]{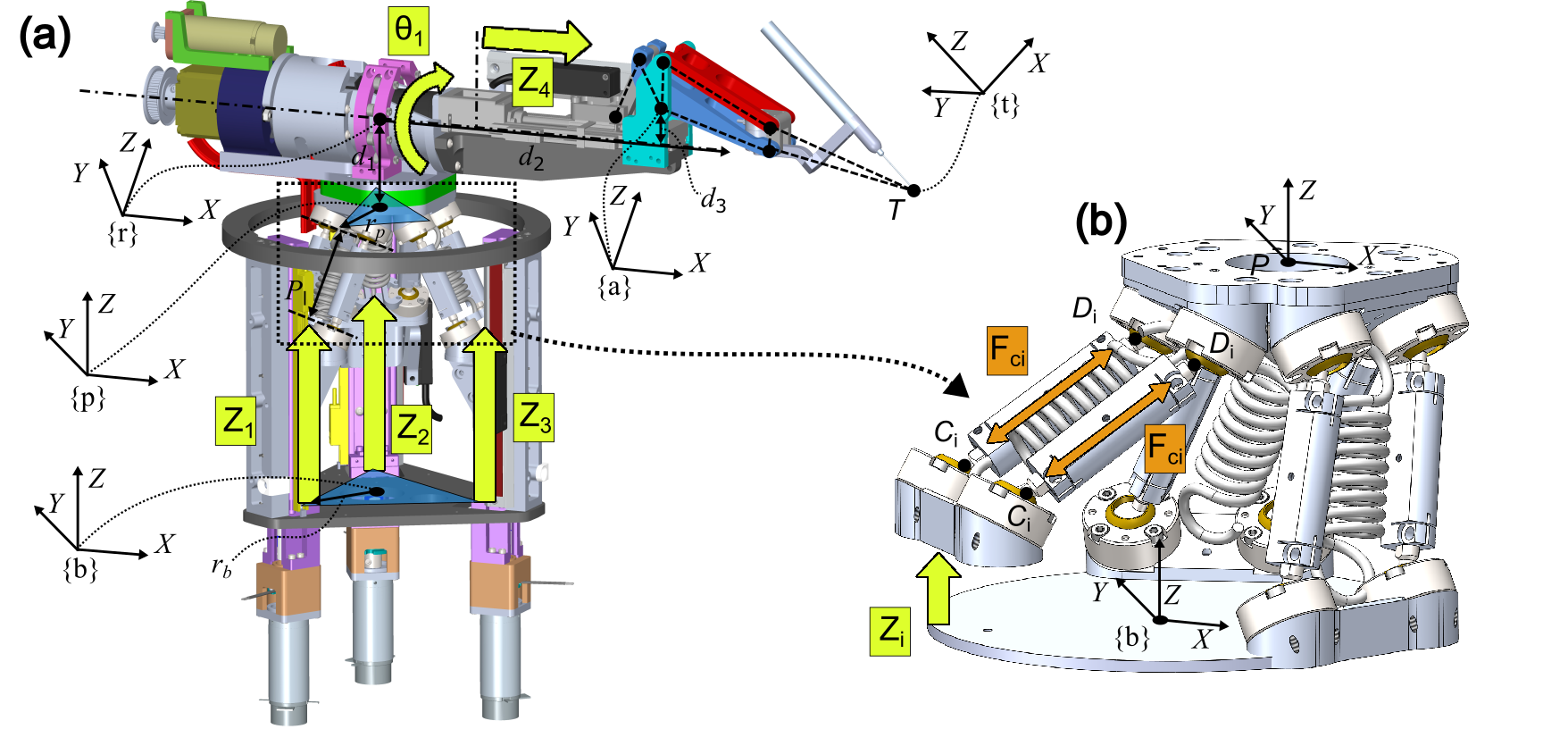}
    \caption{Kinematics diagram of the SHER 3.0. (a) DH Parameters and basic kinematics of the robot arm with the delta stand. Seen are the base frame $\{b\}$, delta platform frame $\{p\}$, roll frame $\{r\}$, four-bar mechanism frame $\{a\}$, and tilt frame $\{t\}$. Kinematics of the arm between frame $\{r\}$ and $\{p\}$ are described more in \cite{wu2020optimized, roth2021towards}. $Z_1$-$Z_4$ represent linear motion for the delta stage and tilt mechanisms and $\theta_1$ represents the rotation about the x-axis in $\{r\}$. (b) Close up of the delta platform showing the offset of a single link $Z_i$. Points $C_i$ and $D_i$ define link $i$'s endpoints and $F_{ci}$ defines the axial force along the link.
    }
    
    \label{fig:kinematicDiagram}
\end{figure*}

Over the last few decades, several robotic systems have been developed for intraocular microsurgery, of which some have been successfully evaluated in clinical trials \cite{de2018robotic, roizenblatt2018robot, vander2020robotic}. Generally, three main approaches have been taken for controlling these robotic systems: (1) telemanipulation, where the system provides tremor filtering and motion scaling (e.g., the PRECEYES Surgical System \cite{de2016robotic} and the Intraocular Robotic Interventional Surgical System (IRISS) \cite{wilson2018intraocular} ); (2) comanipulation (“co-bots”), where the system supports tremor filtering but lacks motion scaling (e.g., the Johns Hopkins Steady-Hand Eye Robot (SHER) \cite{he2012toward} and the KU Leuven robot \cite{gijbels2013design}); and (3) handheld devices, which have minimal impact on surgical workflow (e.g., the Micron handheld tool for microsurgery \cite{yang2014manipulator}).
Of these systems, only the PRECEYES is commercially available in ocular surgery \cite{preceyes}. The system was used in the first successful human intraocular robotic surgery \cite{edwards2018first} and has been used in multiple clinical studies since followed by the KU Leuven robot in a preliminary clinical feasibility study \cite{willekens2017robot}.


\begin{figure}[!t]
    \centering
    \vspace{0.2cm}
    \includegraphics[width=0.7\linewidth]{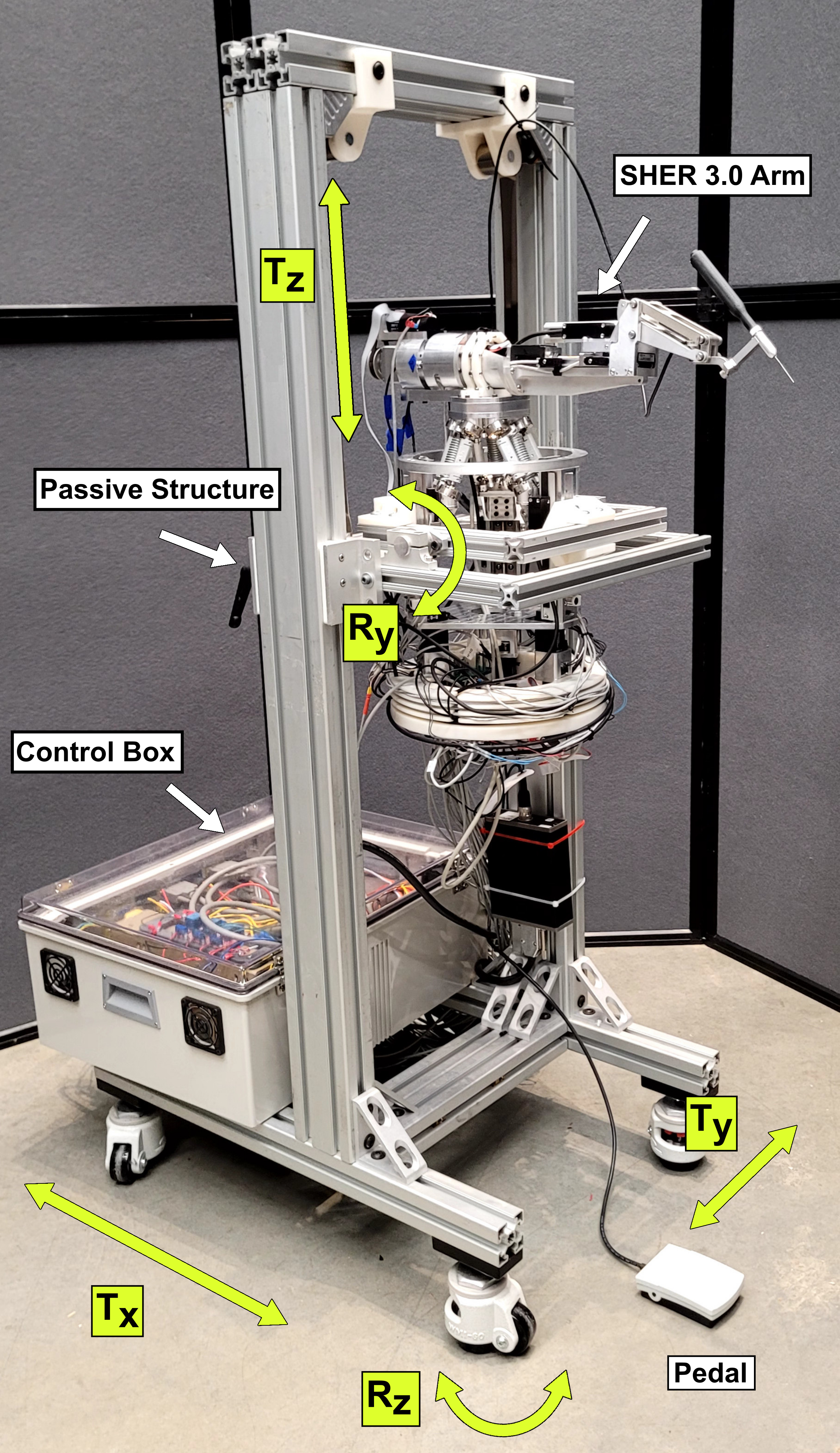}
    \caption{The SHER 3.0 configured on its passive structure with a control box and pedal. The structure allows 5-DOF gross motion to the robot for increased motion and mobility. This includes x and y translation with z-axis rotation on the ground, adjustable z-axis translation for the arm, and a y-axis rotation below the delta stage base for an additional static tilt.}
    \label{fig:full_sher_3}
\end{figure}







Despite the advancements in intraocular robotics, there are currently no robotic platforms capable of supporting safe micromanipulations in the subretinal space which include a delta stage (providing several benefits including a light-weight structure, a compact design, and structural stiffness, among others) 
that would be accepted and trusted by clinicians for future ocular clinical therapeutic tasks (i.e., subretinal injection of stem cell or gene therapies) \cite{iordachita2022robotic, ladha2022subretinal}. This work developed and evaluated a precise and robust robotic platform meeting the clinical requirements for effective subretinal injections. A previous robot design based on \cite{he2012toward, uneri2010new} was updated and improved to do this. This new robot (see Fig. \ref{fig: whole prototype}) is the next-generation Steady-Hand Eye Robot (SHER 3.0). Like previous iterations, it also employs a cooperative control approach, in which both the surgeon and robot hold the surgical tool, and the robot detects forces exerted by the surgeon on the instrument and moves in corresponding directions. Advantages of our new design over other robotic systems include (1) optimized delta stage and arm with high accuracy localization and tremor reduction, (2) direct coupling to human natural motion sensation via cooperative control methods employed, (3) more straightforward integration into existing operating environments \cite{vander2020robotic}, and (4) a lower cost and smaller footprint as compared to previous SHER iterations.
The concept design of the new robotic system is illustrated in Fig. \ref{fig:kinematicDiagram}, and the whole robot system can be seen in Fig. \ref{fig:full_sher_3}. The 5-DOF (degree-of-freedom) robotic system is composed of a 1-DOF roll and 1-DOF tilt mechanism arm \cite{wu2020optimized, roth2021towards} and a 3-DOF linear stage (delta robot) for translational motion. The robotic system is supported by a passive structure that can be nicely situated next to a surgical table.

This work describes the improved SHER 3.0 robot and the design optimization and analysis used to arrive at its final working state. The main contributions of this paper are summarized as follows:
\begin{itemize}
  \item Presentation of the first parallel structure (delta stage) in a robot for cooperatively controlled eye surgery tasks.
  \item The derivation of an accurate deflection model for the delta stage links in preparation for optimization.
  \item An optimization of the delta stage robot design maximizing the global conditioning index and minimizing the average torsional compliance.
  \item Robot accuracy, repeatability, and deflection validation tests.
  
\end{itemize}

This paper is organized as follows. Section \ref{sec:SysDescrip} describes the delta manipulator's design requirements, components, and general high-level kinematics. Section \ref{section:deflection_model} describes the delta stage deflection model derivation and associated tests. This model is used to optimize the delta stage designs. Section \ref{sec:DesignOpt} then explains the parameters, constraints, and optimization objectives used to improve the performance and feasibility of the delta stage. The results from this optimization are then reported. Results from benchtop testing of the robot analyzing accuracy, repeatability, and deflection are noted in section \ref{sec:BenchtopEval}, along with a discussion and ideas for future work.

\section{System Description}\label{sec:SysDescrip}
\subsection{Design Requirements}
One major technical challenge in subretinal injections is the safe delivery of adequate doses of the therapy medium to the target area. In one popular subretinal injection modality, following conventional pars plana vitrectomy, clinicians penetrate the retina with a microcannula and inject precise therapy suspensions into the subretinal space, the area between the internal limiting membrane (ILM) and retinal pigment epithelium (RPE) layers \cite{vander2020robotic}. As the average thickness of the human retina is around 250\,$\mu$m \cite{jo2011diurnal}, the acceptable maximum error for needle tip localization in the vertical direction (from ILM to RPE) is 25\,$\mu$m \cite{zhou2018precision, vander2020robotic}. Based on these standards and our previous experience \cite{he2012toward, uneri2010new} and considering human eye size and specifications for other retinal robotic systems \cite{vander2020robotic}, rotation and translation requirements shown in Table \ref{tab:kinematicSpecs} were proposed for the SHER 3.0 iteration. These requirements outline the range and precision for robot Cartesian translation and X- and Y-rotation (roll and tilt) for the robot arm holding the surgical tool.

\begin{table}[tb]
\centering
\caption{SHER 3.0 kinematic specifications. To visualize associated axes, see Fig. \ref{fig:kinematicDiagram}}
\label{tab:kinematicSpecs}
\begin{tabular}{|lll|}
\hline
DOF                             & Range                & Repeatability   \\ \hline
Rotation $R_x$ in $\{r\}$                      & $\pm 45^\circ$                      & $0.01^\circ$                       \\ 
Rotation $R_y$ in $\{t\}$                     & $\pm 120^\circ$                      & $0.01^\circ$                         \\ 
Translation $T_x$ in $\{p\}$                      & $\pm 27.5mm$                      &  $5\mu m$               \\ 
Translation $T_y$ in $\{p\}$                     & $\pm 27.5mm$                      &  $5\mu m$    \\ 
Translation $T_z$ in $\{p\}$                     & $\pm 30mm$                      &  $5\mu m$    \\ \hline
\end{tabular}%
\label{tab:COM_pos_table}
\end{table}

Table \ref{tab:kinematicSpecs} shows preliminary system specifications set as: (1) $R_x$ motion, ±45º range, 0.01º precision, (2) $R_y$ motion, ±120º range, 0.01º precision, (3) $T_x$, $T_y$,  ±27.5\,mm (55\,mm total), and $T_z$ motions, ±30\,mm (60\,mm total) range, 5\,$\mu$m precision. The most vital direction for safe subretinal injection is the axis perpendicular to the retina face as it will be touched by the tip at $\{t\}$. This is because the puncture distance must be accurate enough to inject the therapy for bleb formation between the appropriate retinal layers. Our goal was to achieve a tip accuracy of 25-30\,$\mu$m  as defined by evaluations for subretinal injections from our clinician partners and recent results achieved by related eye surgery robots \cite{wang20225}. 

\subsection{Conceptual Design}

The robot design can be seen in Fig. \ref{fig: whole prototype}, \ref{fig:kinematicDiagram}, and \ref{fig:full_sher_3}. The base of SHER 3.0 uses a delta mechanism to provide linear Cartesian motion. Compared to the Cartesian stages used in previous iterations of the SHER \cite{uneri2010new, Mitchell2007}, this new delta mechanism improves the positioning accuracy and rigidity of the entire system while also reducing the robot geometrical envelope size.
Three DC motors (RE 25, Maxon Motors) drive the movement of the slider on the high precision linear stages (KR Direct Motor Mount, THK), while six links forming three legs are connected between the three sliders and the upper base via ball joints (SR joints, Hephaist) to create a parallel delta mechanism.
The ball joints were selected to improve positioning accuracy with a backlash of just $\pm$ 1\,$\mu$m.

An arm is mounted on the delta stage with a roll and tilt mechanism. The roll mechanism contains a motor (RE 25, Maxon Motors) connected to a 1/100 harmonic drive gearbox (CSF-11-100-1U-F, Harmonic Drive Systems Inc.) via a timing belt mechanism. Using a harmonic reducer reduces the clearance of the drive mechanism and thus improves the accuracy of the rotary joint. The roll apparatus is followed by the tilt mechanism described in detail in our previous papers \cite{wu2020optimized,roth2021towards}. The tilt mechanism uses a four-bar linkage scheme designed so it can be used on either side of the patient according to surgeon preference, surgeon hand dominance, and operating room space constraints. Optimizing the mechanism minimizes the working space required to ensure compensatory movements of the virtual remote center of motion. Linear absolute encoders (magnetic coding disk (AskIM-2, Renishaw) for the roll joint, and linear absolute linear encoders (LA11, Renishaw) for all other joints)  were connected to all actuators which record the position of the robot joints, reducing the need for homing when power-cycled. An ophthalmic instrument is rigidly connected to a force sensor (Nano17, ATI Industrial Automation) near the handle so that the force generated by the surgeon while manipulating the instrument can be used to create the robot's synergistic motion.

The delta stage and arm were mounted onto a passive structure seen in Fig. \ref{fig:full_sher_3}. The structure allows 5 DOF static motion in addition to the 5-DOF of the robot to enable better mobility and adaptation to the surgical environment. It is paired with a gas pedal and control box for using the cooperative-control features of the robot.

\subsection{Kinematics}

The SHER 3.0 works cooperatively with the surgeon in an admittance control mode. In this mode, the robot aims to create a velocity proportional to detected handle forces in task space which are applied by the surgeon holding the attached tool. The proportionality of this relationship was tuned based on feedback from experienced users and clinicians. The forward kinematics of the whole robot can be described as

\begin{equation}
    \prescript{\{ b \}}{}{\textbf{T}}_{\{ t \}} =  \prescript{\{ b \}}{}{\textbf{T}}_{\{ p \}}  \prescript{\{ p \}}{}{\textbf{T}}_{\{ r \}}  \prescript{\{ r \}}{}{\textbf{T}}_{\{ a \}}  \prescript{\{ a \}}{}{\textbf{T}}_{\{ t \}}
\end{equation}
where $\{b\}$, $\{p\}$, $\{r\}$, $\{a\}$, and $\{t\}$ describe the base,  platform, roll, four-bar mechanism, and tip frames respectively (see Fig. \ref{fig:kinematicDiagram}).
$\prescript{\{ b \}}{}{\textbf{T}}_{\{ p \}}$ and $\prescript{\{ a \}}{}{\textbf{T}}_{\{ t \}}$ are obtained from the forward kinematics of the delta mechanism and tilting mechanism respectively, and details surrounding kinematic computation and frames are described more extensively in \cite{wu2020optimized} and \cite{roth2021towards}. $\prescript{\{ p \}}{}{\textbf{T}}_{\{ r \}}$ and $\prescript{\{ r \}}{}{\textbf{T}}_{\{ a \}}$ can be described as

\[\prescript{\{ p \}}{}{\textbf{T}}_{\{ r \}} = trans(Z, d_1)rot(X, \theta_1)\]
\[ \prescript{\{ r \}}{}{\textbf{T}}_{\{ a \}} = trans(X, d_2)trans(Z, d_3)\]

where $trans$ and $rot$ denote translation and rotation about the given axis.

\section{Delta Stage Deflection Model} \label{section:deflection_model}


In preparation for performing an optimization of delta stage designs to improve robot torsional compliance, a proper deflection model is needed. This section describes the developed deflection model for the delta mechanism to determine the torsional compliance for each design configuration. This model was adopted from a similar work by Ahamad et al. \cite{ahmad2014stiffness}.
Its purpose is to find the linear and angular deflection of the delta robot end-effector point $P$ (denoted by $[\delta \boldsymbol{X}_P \; \delta \boldsymbol{\theta}_P]^T$) for a given external wrench at the same point ($[\boldsymbol{f}_{P} \; \boldsymbol{\tau}_{P}]^T$).

\subsection{Method}

To develop the compliance model for the delta stage, the following assumptions are made (see Fig. \ref{fig:kinematicDiagram} for reference):

\begin{itemize}
    \item The base $\{b\}$ and moving $\{p\}$ platforms are rigid and the only source of deflection is the links' extension and compression.
    \item The SR joints are frictionless so the links only pass tension or compression forces along their axes; in other words, the torques $\boldsymbol{\tau}_{C_i} = 0$ and $\boldsymbol{\tau}_{D_i} = 0$, and the axial force $F_{ci}$ is in the direction of the line connecting $C_i$ and $D_i$ ($\boldsymbol{r}_{C_i D_i}$).
    \item Any deflection resulting from an external force is considered an infinitesimal motion.
    \item The combined axial compliance of the delta links and SR joints is nonlinear and a function of axial load (expressed as $\mathcal{C}(F_{c})$). $\mathcal{C}(F_{c})$ is identified through experiment and discussed later in this section.
\end{itemize}

For a given external wrench at $P$, the resultant wrench at point $D_i$ (refer to Fig. \ref{fig:kinematicDiagram}) can be calculated from

\begin{equation}
    \begin{bmatrix}
        \boldsymbol{f}_{P} \\ \boldsymbol{\tau}_{P}
    \end{bmatrix} = \sum_{i=1}^{6}
    \begin{bmatrix}
        \textbf{I} & \textbf{0} \\ [\boldsymbol{r}_{PD_i}\times] & \textbf{I}
    \end{bmatrix}
    \begin{bmatrix}
        \boldsymbol{f}_{D_i} \\ \boldsymbol{\tau}_{D_i}
    \end{bmatrix}
    \label{eq:compute_wrench_p_from_d}
\end{equation}

where $[\boldsymbol{r}_{PD_i}\times]$ is the skew-symmetric matrix of vector $\boldsymbol{r}_{PD_i}$. Since $\boldsymbol{\tau}_{D_i} = \boldsymbol{0}$ and $\boldsymbol{f}_{D_i} = [0\;0\;F_{ci}]^T$ equation \ref{eq:compute_wrench_p_from_d} reduces to

\begin{equation}\label{eq:compute_wrench_p_from_fci_vector_form}
    \begin{bmatrix}
        \boldsymbol{f}_{P} \\ \boldsymbol{\tau}_{P}
    \end{bmatrix} = \frac{1}{L}\sum_{i=1}^{6}
    \boldsymbol{\gamma}_i  F_{ci}
\end{equation}

where $\boldsymbol{\gamma}_i$ is defined as  

\begin{equation}\label{eq:gamma}
    \boldsymbol{\gamma}_i=
    \begin{bmatrix}
        \boldsymbol{r}_{C_i D_i} \\ \boldsymbol{r}_{PD_i} \times \boldsymbol{r}_{C_i D_i}
    \end{bmatrix}_{6\times1}.
\end{equation}

Equation \ref{eq:compute_wrench_p_from_fci_vector_form} can be rewritten in matrix form as

\begin{equation}\label{eq:compute_wrench_p_from_fci_matrix_form}
    \begin{bmatrix}
        \boldsymbol{f}_{P} \\ \boldsymbol{\tau}_{P}
    \end{bmatrix} = \frac{1}{L} \boldsymbol{\Gamma}  \boldsymbol{F}_{c}
\end{equation}

where 

\[   \boldsymbol{\Gamma} = \begin{bmatrix}      
    \boldsymbol{\gamma}_1 & \dots & \boldsymbol{\gamma}_6 \end{bmatrix} 
\]

and 

\[ \boldsymbol{F}_{c} = \begin{bmatrix} F_{c1} & \dots & F_{c6} \end{bmatrix}^T. \]

So the resultant axial forces of the delta links for the given external wrench at point $P$ are obtained as

\begin{equation}\label{eq:compute_fci_from_wrench_p_matrix_form}
     \boldsymbol{F}_{c} = L \boldsymbol{\Gamma}^{-1} \begin{bmatrix}
        \boldsymbol{f}_{P} \\ \boldsymbol{\tau}_{P}.
    \end{bmatrix}
\end{equation}

The next step is to establish the deflection behavior of the delta platform. Since point $C_i$ (refer to Fig. \ref{fig:kinematicDiagram}(b)) is stationary, the small deflection of each link $\delta L_i$ can be expressed as the infinitesimal displacement of point $D_i$ along link axis $\boldsymbol{r}_{C_i D_i}$, which is 

\begin{equation}\label{eq:delta_Li}
\delta L_i = \frac{\boldsymbol{r}_{C_i D_i} \cdot \delta \boldsymbol{X}_{D_i}}{L}.
\end{equation}

$\delta \boldsymbol{X}_{D_i}$ is the displacement of point $D_i$ and is related to the infinitesimal motion of the delta end-effector point $P$ as

\begin{equation}\label{eq:delta_X_Di}
\delta \boldsymbol{X}_{D_i}=\delta \boldsymbol{X}_P + \delta \boldsymbol{\theta}_P\times \boldsymbol{r}_{P D_i}
\end{equation}

 where $\delta\boldsymbol{X}_P$ and $\delta\boldsymbol{\theta}_P$ are the infinitesimal linear and angular motions of point $P$, respectively.
 
Plugging \ref{eq:delta_X_Di} into \ref{eq:delta_Li} and then using the cyclic property of the scalar triple product results in 

\begin{equation}\label{eq:delta_Li_rearranged}
\delta L_i = \frac{1}{L} \{ \boldsymbol{r}_{C_i D_i} \cdot \delta \boldsymbol{X}_P + (\boldsymbol{r}_{P D_i} \times \boldsymbol{r}_{C_i D_i})\cdot \delta \boldsymbol{\theta}_P\}
\end{equation}

which can be rewritten as

\begin{equation}\label{eq:delta_Li_from_P_deflection}
    \delta L_i = \frac{1}{L}
    \boldsymbol{\gamma}_i^{T} 
    \begin{bmatrix}
        \delta \boldsymbol{X}_P \\ \delta \boldsymbol{\theta}_P
    \end{bmatrix}
\end{equation}

and then in matrix form as

\begin{equation}\label{eq:delta_L_from_P_deflection}
    \delta \boldsymbol{L} = \frac{\boldsymbol{\Gamma}^{T}}{L}
    \begin{bmatrix}
        \delta \boldsymbol{X}_P \\ \delta \boldsymbol{\theta}_P
    \end{bmatrix}
\end{equation}

where 
\[ \delta \boldsymbol{L} = \begin{bmatrix} \delta L_1 & \dots & \delta L_6 \end{bmatrix}^T. \]

Finally, deflection of point $P$ can be expressed as

\begin{equation}\label{eq:P_deflection_from_delta_L}
    \begin{bmatrix}
        \delta \boldsymbol{X}_P \\ \delta \boldsymbol{\theta}_P
    \end{bmatrix} = 
    L \{\boldsymbol{\Gamma}^{T}\}^{-1} \delta \boldsymbol{L}.
\end{equation}

To complete the model, we need the relationship between $\boldsymbol{F}_{c}$ and $\delta \boldsymbol{L}$, which can be expressed as

\begin{equation}\label{eq:compliance_of_link}
    \delta \boldsymbol{L} = \mathcal{C}(\boldsymbol{F}_{c})\boldsymbol{F}_{c}
\end{equation}

where $\mathcal{C}(\boldsymbol{F}_{c})$ is the nonlinear compliance of the delta links and is identified through several experiments presented hereafter.

In summary, equations \ref{eq:compute_fci_from_wrench_p_matrix_form}, \ref{eq:P_deflection_from_delta_L}, and \ref{eq:compliance_of_link} describe the analytical model used in section \ref{sec:DesignOpt} to calculate the torsional compliance of the delta platform in different design configurations.


\subsection{Experimental Setup}

\begin{figure}
	\centerline{\includegraphics[width=3.5in]{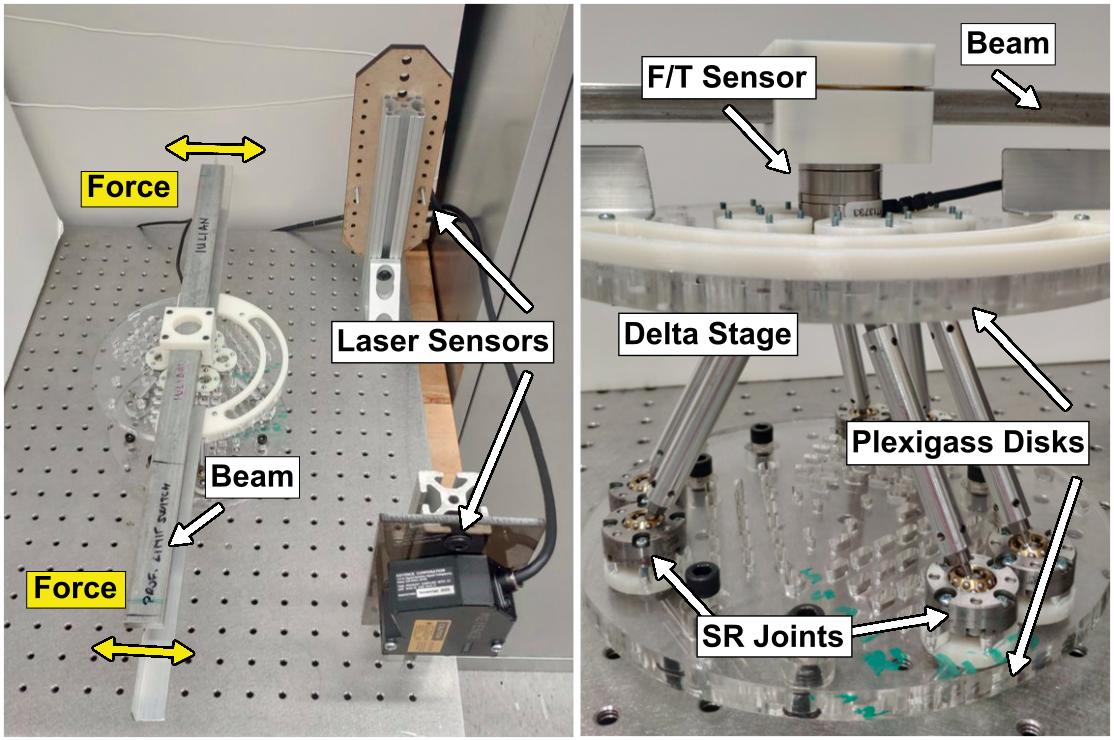}}
	\caption{Test setup for measuring delta torsional deflection for different parameters.}
	\label{fig:deflection_test_setup}
\end{figure}

A series of tests were performed to verify the deflection model and identify the compliance of the delta stage links, $\mathcal{C}$. Two Plexiglass disks with radially and laterally distributed holes, were used to achieve a variety of base radius sizes $r_b$, moving platform radius sizes $r_p$, and leg widths $w$ (refer to Fig. \ref{fig:deflection_test_setup} and \ref{fig_delta_before_after_calibration}(a)). A 6-axis force-torque sensor (Nano25, ATI Industrial Automation) and an aluminum beam were attached to the center of the moving platform $P$ to exert and measure a torsional moment at point $P$ as seen in Fig. \ref{fig:deflection_test_setup}. The capacity and resolution of the sensor for its torsional moment $\tau_z$ were $3.4$\,Nm and $1/1320$\,Nm respectively. Two laser sensors (LK-H157, Keyence Corporation of America), with an active range of $150\pm$40\,mm and linearity of $\pm 1.6$\,$\mu$m, were used to read the linear displacement of the ends of the attached beam. These measurements were used to calculate the torsional deflection of the delta platform.

\begin{figure*}[t!]
	\centerline{\includegraphics[width=\textwidth]{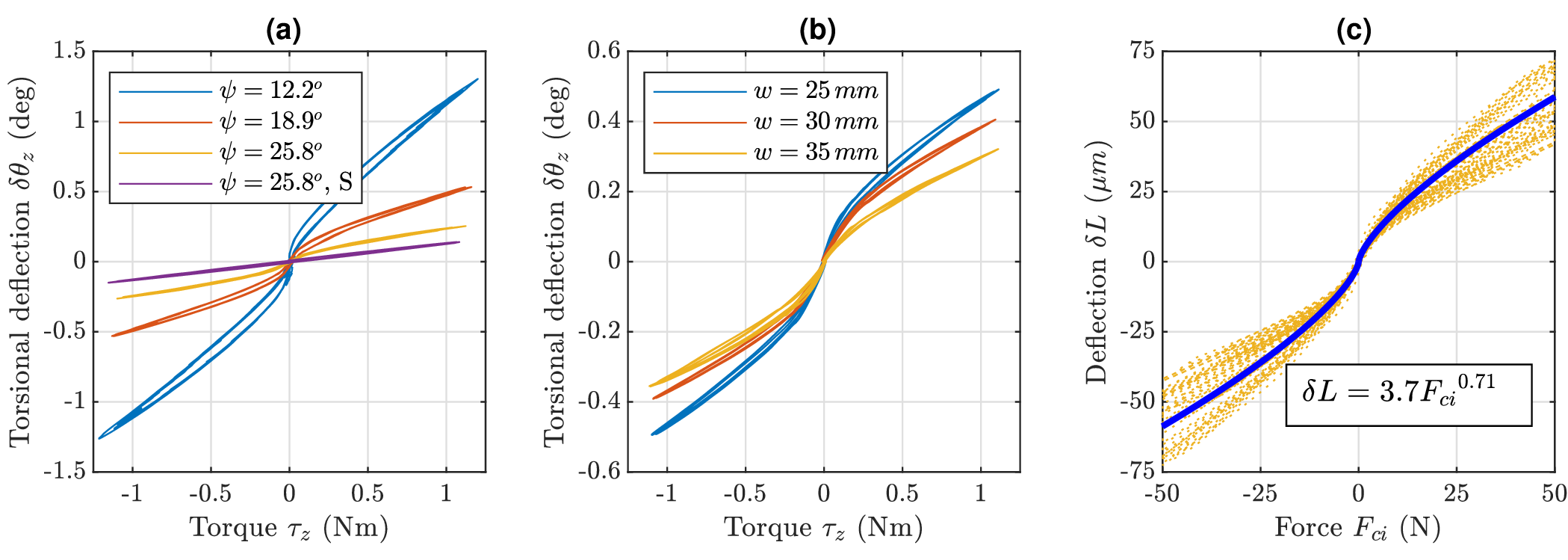}}
	\caption{(a) Torsional deflection vs torsional moment for $r_p=28.8$\,mm, $w=30$\,mm, and different offset angles $\psi$ ($S$ means with spring). (b) Torsional deflection vs torsional moment for $r_p=37.7$\,mm, $\psi=18.9^\circ$, and different leg widths $w$ (c) Deflection vs. force behavior of the delta links for various experiment trials. The calculated deflection behavior for each of the 23 experiments is shown as dashed orange lines and the solid blue line is the fitted curve for these experiments.}
	\label{Fig.MI_result}
\end{figure*}

A total of 25 different configurations were tested. Of these cases, 23 used different geometric parameters for $r_b$, $r_p$, and $w$. The other two instances added springs to the delta stage to study the effect of preloading the SR joints. For each configuration, the user manually exerted increasing and decreasing torsional moments by applying symmetric but opposing forces in both directions at the two ends of the bar. This was repeated for five cycles. The force sensor recorded the forces and torques exerted at point $P$ and the laser sensors recorded the displacements of the two ends of the beam from which the torsional deflection of point $P$ could be computed.

\begin{table}[t!]
    \caption{Average torsional compliance of the delta stage (in deg/Nm) for different geometric parameters measured from the experiment. "S" in the last row means the test included a preloaded spring.}
    \label{tab:compliance_tests}
    \begin{center}
        \begin{tabular}{|m{0.8cm}|m{0.4cm}m{0.4cm}m{0.4cm}|m{0.4cm}m{0.4cm}m{0.4cm}|m{0.4cm}m{0.4cm}m{0.4cm}|}
            \hline
             -  & \multicolumn{3}{c|}{$r_p = 28.8$\,mm} & \multicolumn{3}{c|}{$r_p = 37.7$\,mm} & \multicolumn{3}{c|}{$r_p = 47.0$\,mm} \\
            \hline
            \diagbox[height=1.5\line,innerwidth=0.8cm]{$\psi$}{$w$} & 25 & 30 & 35         & 25 & 30 & 35                       & 25 & 30 & 35 \\
            \hline
            $5.91^\circ$  & - & - & - & - & 3.7 & 3.0 & 4.1 & 3.4 & 2.8 \\
            $12.2^\circ$  & 1.33 & 1.19 & 0.93 & 1.16 & 0.84 & 0.79 & 1.07 & 0.85 & 0.69 \\
            $18.9^\circ$  & 0.64 & 0.53 & 0.41 & 0.52 & 0.43 & 0.35 & - & - & - \\
            $25.8^\circ$  & 0.32 & 0.27 & 0.23 & - & - & - & - & - & - \\
            $25.8^\circ\,S$  & - & 0.13 & - & - & - & - & - & - & - \\
            \hline
        \end{tabular}
    \end{center}
\end{table}

The results of these experiments were used to identify $\mathcal{C}$. For each test, the torsional torque data $\tau_z$ was inserted into equation \ref{eq:compute_fci_from_wrench_p_matrix_form} and corresponding link forces $\boldsymbol{F}_c$ were calculated. At the same time, the torsional deflection data $\delta\theta_z$ was given to equation \ref{eq:delta_L_from_P_deflection} and resultant link deflections $\delta\boldsymbol{L}$ were determined. The relationship between the link forces and link deflections were investigated, resulting in a model for compliance behavior of the link, $\mathcal{C}$.


\subsection{Results and Discussion}

Table \ref{tab:compliance_tests} shows the average torsional compliance of the delta for different geometric parameters. Fig. \ref{Fig.MI_result}(a) and \ref{Fig.MI_result}(b) depict the relationship between the external torsion $\tau_z$ and the corresponding torsional deflection $\delta \theta_z$ for a select number of experiments. Fig. \ref{Fig.MI_result}(a) corresponds to $w=30$\,mm, $r_p=28.8$\,mm, and different offset angles $\psi$.  It also includes one test in which tension springs were added between each link pair, as seen in Fig. \ref{fig_delta_before_after_calibration}(b). Fig. \ref{Fig.MI_result}(b) shows the effect of different leg widths $w$ (distance between link pairs), for $r_p=37.7mm$, $\psi=18.9^\circ$, on torsional deflection.

It can be seen that increasing the offset angle $\psi$ decreased the torsional compliance thus improving the torsional rigidity of the delta platform. Preloading the SR joints by a tension spring also increased torsional rigidity. These results align with our initial expectation that preloading the SR joints would eliminate any backlash or clearance between the SR joint components (bowl, balls, and socket) and in turn improve rigidity. Fig. \ref{Fig.MI_result}(b) shows that increasing the leg width $w$ increased torsional rigidity, and Table \ref{tab:compliance_tests} shows increasing $r_p$ also slightly decreased the compliance for a given $\psi$ and $w$. Both behaviors were expected under the assumption that the compliance of the delta stage is mainly caused by the compliance of the delta links.

Finally, Fig. \ref{Fig.MI_result}(c) shows the resultant relationship between the axial force and axial deflection of delta links after plugging the experimental data in the analytical model (equations \ref{eq:compute_fci_from_wrench_p_matrix_form} and \ref{eq:delta_L_from_P_deflection}). For an ideal model and ideal experiment, we expect to see similar outcome for all the tests. But we are using a simple model that assumes the only source of deflection in delta platform is the deflection of the links and that all the links are identical. Additionally, the experimental setup and data acquisition introduce some errors. In Fig. \ref{Fig.MI_result}(c) curve fitting was performed using a power equation resulting in  $\delta L = 3.7 {F_{c}}^{0.71}$, where $\delta L$ is in $\mu$m and $F_{c}$ is in $N$. For the fitted curve, the R-squared value is $0.91$ and the RMS of error is $5.25$\,$\mu$m.

\section{Design Optimization}\label{sec:DesignOpt}

Optimization of the tilting mechanism is extensively discussed in \cite{roth2021towards}. The optimization conducted here focuses on optimizing the delta mechanism for improving the larger motion of the robot. The first prototype of the delta robot for the SHER 3.0 used the Global Conditioning Index (GCI) only to define the optimization goal. GCI is a well-known kinematic performance criterion defined as the average of the inverse of the condition number, i.e. the ratio between the largest and smallest eigenvalues of the Jacobian matrix throughout the workspace. Its value ranges between zero and one. A larger GCI denotes a better share of actuators in creating a given motion at the end-effector and thus better kinematic performance.
After building the first prototype, we observed unacceptable deflection at the robot's end-effector, mainly due to angular compliance of the delta robot. As shown in Fig. \ref{fig:kinematicDiagram}, the relatively long arm attached to the delta stage could allow even a small force at the end-effector $\{t\}$ to create a large moment at the moving platform frame origin $\{p\}$. 

To address this problem, the optimization was performed again, however this time factoring angular compliance based on the developed deflection model into the optimization in addition to GCI. This model was then used to calculate the angular compliance of the delta stage in each of the design configurations. 

\subsection{Problem definition}

The four factors influencing optimization were design parameters, optimization constraints, the GCI, and angular compliance. For each set of design parameters an exhaustive search was performed through the target workspace. If all constraints were met the average of GCI and angular compliance was calculated throughout the workspace. Lower compliance and GCI closer to one were preferred. The optimization details including the parameters, constraints, and objectives are explained next.

\subsubsection{Design parameters}
Fig. \ref{fig_delta_before_after_calibration}(a) shows the kinematic independent optimization parameters for the delta robot. These were link length $L$, leg width (distance between each link pair in a leg) $w$, and leg base offset angle (offset angle) $\psi$. The ranges and step sizes for these parameters were chosen as $L$(60:2:90\,mm), $w$(25:5:40\,mm), and $\psi$(18:1:32\,deg). The radius of the moving platform $r_p$ is calculated based on the radius of the spherical joints $r_{SR}$, and the leg width $w$. It can be defined as the minimum radius of the moving platform that can accommodate all six spherical joints and is calculated as

\begin{equation}\label{eq:rp}
r_p=\frac{\sqrt{4r_{SR}^{2} + 2wr_{SR} + w^2}}{\sqrt{3}}.
\end{equation}

The radius of the base platform, $r_b$ can be then calculated as \(r_b = r_p+L sin(\psi)\).

\subsubsection{Optimization constraints}
After several design iterations, we found the following set of constraints to be effective for optimization improvements:

\begin{itemize}
    \item The center of the moving platform should maintain a cylindrical workspace with a diameter of 55\,mm and height of 60\,mm as defined in Table \ref{tab:kinematicSpecs}.
    \item No collision should occur between the moving platform and the outer diameter ring extending vertically from the base platform ($r_b - r_p > 30$\,mm) (see Fig. \ref{fig:kinematicDiagram}).
    \item The prismatic joint  linear motion (i.e. the motion from the three delta linear stages) maximum range should be $<$ 90\,mm.
    \item The maximum swing angle of the spherical joints should be less than $\pm$30\,deg.
\end{itemize}

\subsubsection{Optimization Objective 1 -- Maximize Global Conditioning Index}
As explained previously, the GCI, a well-known kinematic performance criterion, was used as part of the optimization criterion. A larger GCI denoted a better share of actuators and kinematic performance and was preferred.

\subsubsection{Optimization Objective 2 -- Minimize Torsional Compliance}

Finding a design to minimize average torsional compliance $\widehat{T_c}$ (defined in equation \ref{eq:avg_tors_comp}) over the tested configurations was desirable to increase robot stiffness.

Here we used the analytical model explained in section \ref{section:deflection_model} to  determine the torsional compliance for each design. 
Torsional compliance is pose dependent and thus is different for varying configurations of the end-effector $P$. For each configuration point $P$ analyzed in the delta stage workspace, torsional compliance was computed and averaged over all configuration points found. This average score $\widehat{T_c}$ became the objective function. A design with smaller $\widehat{T_c}$ was preferred.

For each design (set of optimization parameters $L$, $w$, and $\psi$) for a given external torsional torque at point $P$ ($\boldsymbol{f}_P = [0\; 0\; 0]^T$ and $\boldsymbol{\tau}_P = [0\; 0 \; \tau_z]^T$), equation \ref{eq:compute_fci_from_wrench_p_matrix_form} determines the axial forces of the delta links $\boldsymbol{F}_{ci}$, equation \ref{eq:compliance_of_link} calculates the resultant deflection of each link.Finally, equation \ref{eq:P_deflection_from_delta_L} provides the resultant deflection of point $P$ ($[\delta \boldsymbol{X}_P \; \delta \boldsymbol{\theta}_P]^T$). The torsional deflection $\delta\theta_{Pz}$ was used to compute torsional compliance $T_c$ as

\begin{equation}
    T_c = \frac{\delta\theta_{Pz}}{\tau_z}
\end{equation}

which can be averaged through the delta robot workspace to get $\widehat{T_c}$ as

\begin{equation}\label{eq:avg_tors_comp}
    \widehat{T_c} = \frac{1}{n}\sum_{i=1}^{n}T_c(i).
\end{equation}

for some $n$ number of configurations tested.

\begin{figure}
	\centerline{\includegraphics[width=3.5in]{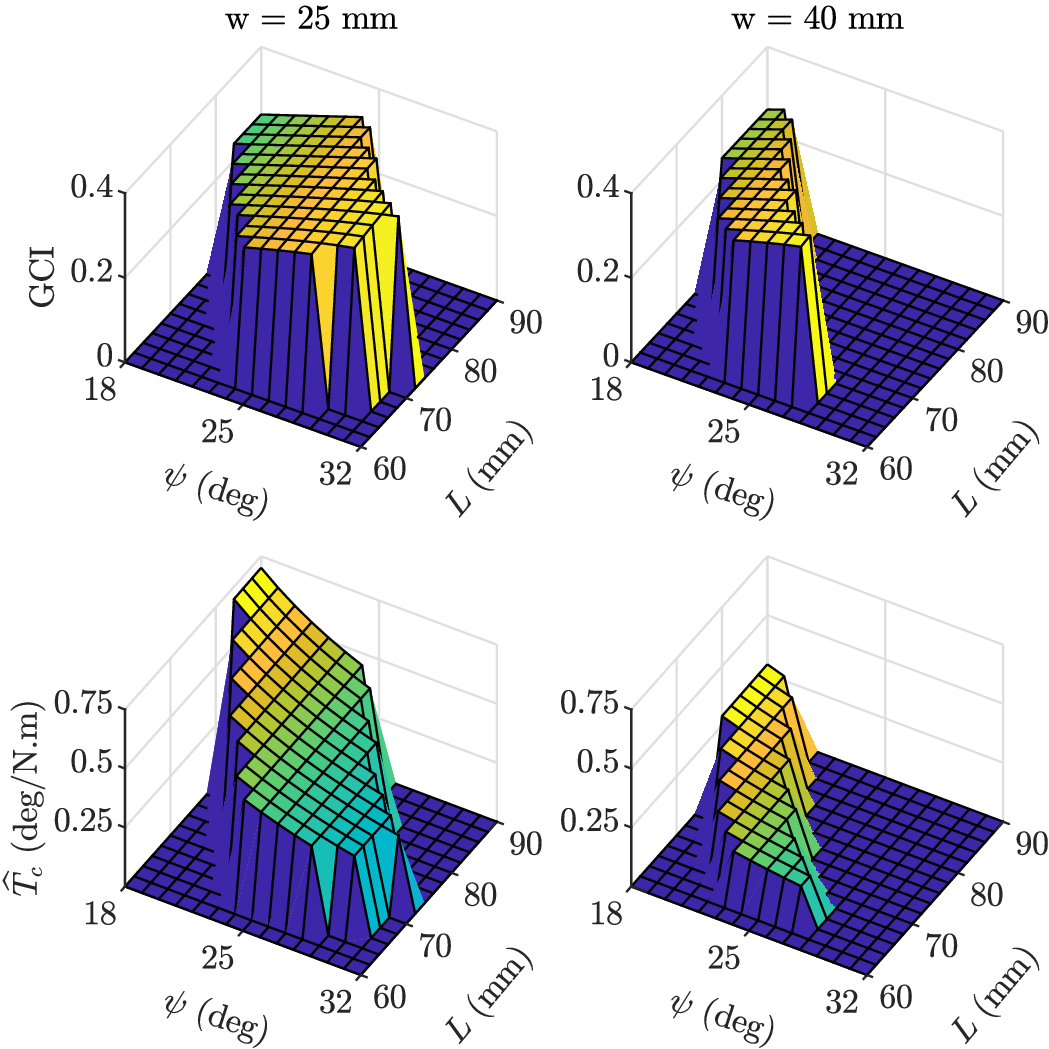}}
	\caption{GCI and average torsional compliance, $\widehat{T_c}$, optimization results for the defined design parameters. Increasing the offset angle, $\psi$, increases the GCI and decreases $\widehat{T_c}$ for the width between link pairs, $w$, values shown. Varying link length, $L$, has a minimal effect on optimization outputs.}
	\label{Fig.opt_results}
\end{figure}

\subsection{Optimization Results}
\begin{figure}
 \centering
 \includegraphics[width=3.5in]{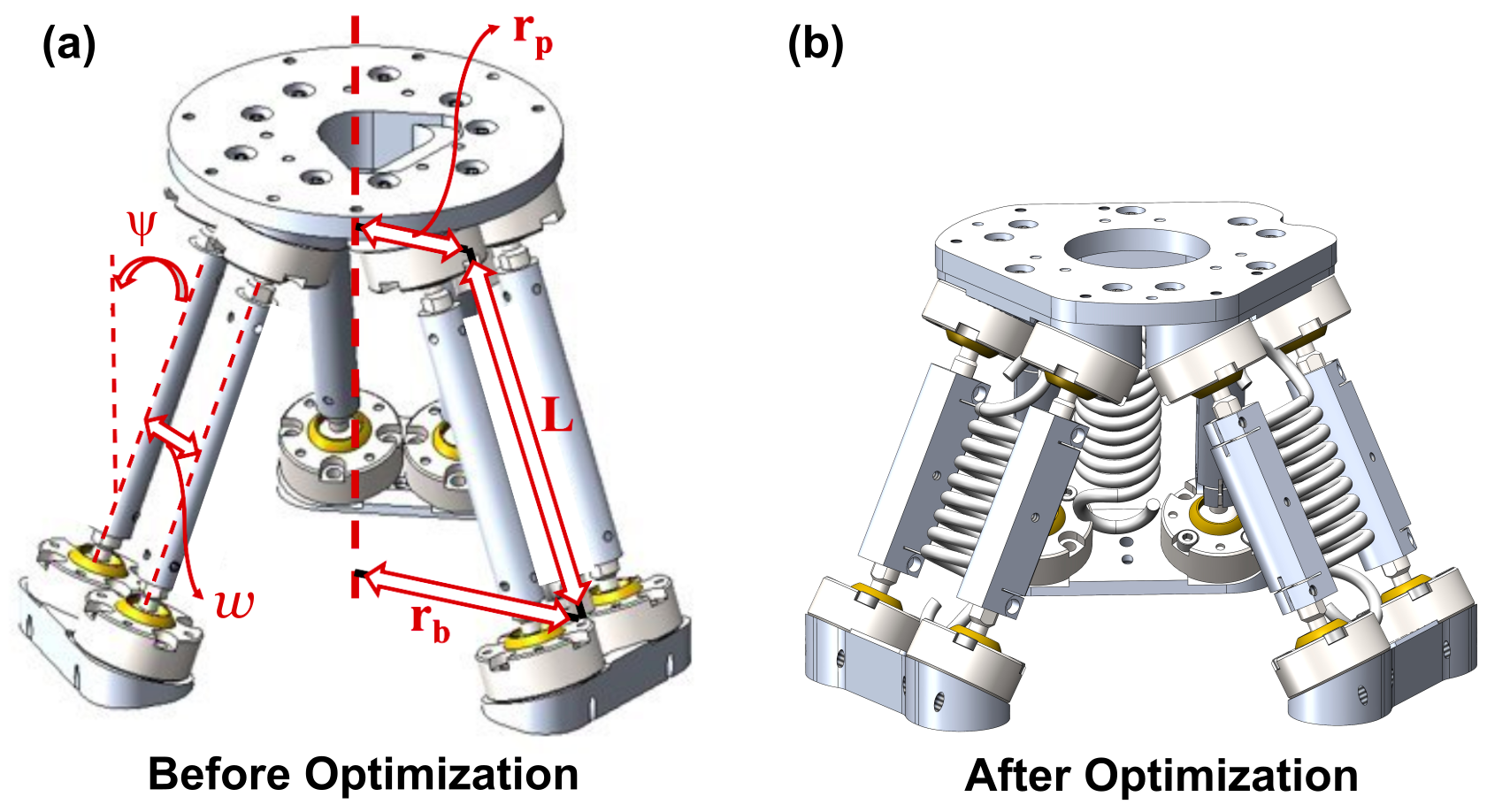}
 \caption{ The optimized delta stage design. (a) Delta stage design before optimization and testing. (b) Updated delta design after optimization. The optimized design is more compact and incorporates springs into the legs to add additional torsional stiffness by pre-compressing the SR joints. }\label{fig_delta_before_after_calibration}
\end{figure}

Fig. \ref{Fig.opt_results} shows the optimization outputs for the GCI and $\widehat{T_c}$ for the highest and lowest values of the parameter $w$ at 25\,mm and 40\,mm. Increasing $\psi$ proved to increase GCI and greatly decrease $\widehat{T_c}$. Leg length $L$ does not seem to play a highly significant role in influencing GCI or $\widehat{T_c}$ within the design parameters. The final parameters selected from this optimization were $w$: $40$\,mm, $L$: $64$\,mm, $\psi$: $27^\circ$, GCI: $0.356$, $\widehat{T_c}$: $0.156$\,deg/Nm, $r_b$: $61.8$\,mm, and $r_p$: $32.8$\,mm.
These parameters were incorporated into a new delta design in Fig. \ref{fig_delta_before_after_calibration}(b) versus the previous design in Fig. \ref{fig_delta_before_after_calibration}(a). The actual parameters before and after the optimization are displayed in Table \ref{tab:optimization_params}

\section{Bench-top Evaluation of Robot Performance}\label{sec:BenchtopEval}

Following the optimization, various performance metrics of the SHER 3.0 were evaluated including accuracy, repeatability, and deflection at individual joints and the needle tip.

\subsection{Experimental Setup}
\begin{figure*}
	\centerline{\includegraphics[width=7.16in]{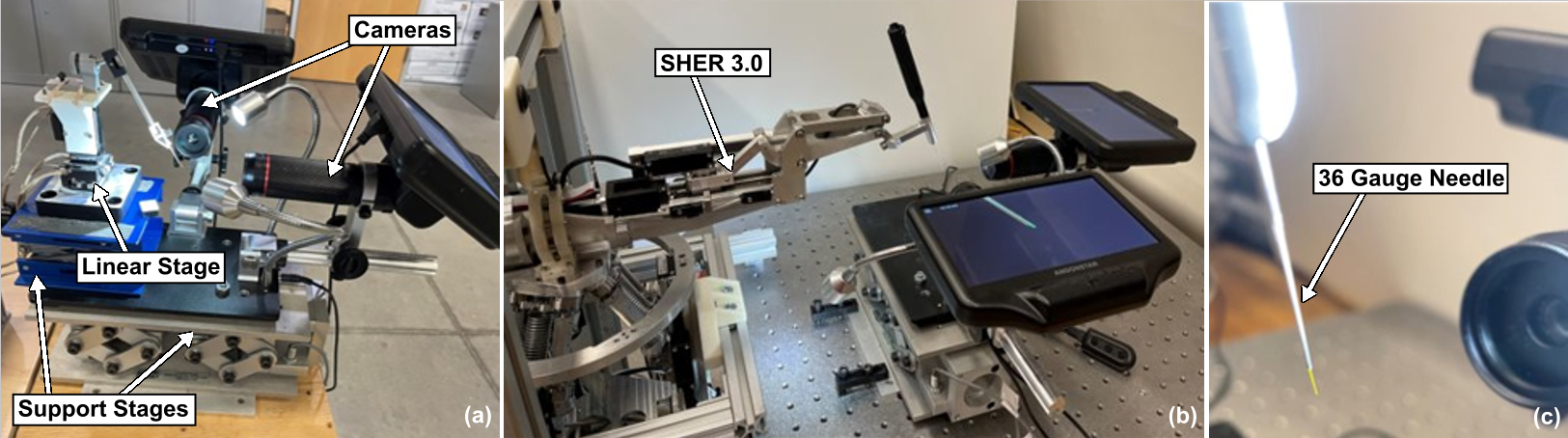}}
	\caption{(a) A Cartesian stage positioning system measures a surgical needle with two sub-micron resolution cameras. (b) The dual camera system is again used, except to measure the needle mounted on the robot arm. (c) A 36 gauge needle is used for camera measurement.}
	\label{fig:camera_setup}
\end{figure*}

Two high resolution digital microscope cameras (AD407 Andonstar, 12M) were used to record the motion of the needle tip in different experiments as seen in Fig. \ref{fig:camera_setup}. To verify the resolution and accuracy of the cameras  we used an xyz linear stage  (Q-522 Q-motion XYZ linear stage, Physik Instrumente (PI)) with sub-micron accuracy as a ground truth to test precise movements.
The cameras were adjusted to change the field of view between either a 10$\times$10\,mm or 2$\times$2\,mm field of view. A camera measurement precision test was conducted to verify the viewing workspace. 
Pixel size and RMS error was 0.8\,$\mu$m and 4.9\,mm respectively for the 10$\times$10\,mm view, and 3.8\,$\mu$m and 18.4\,mm respectively for the 2$\times$2\,mm view.


\begin{table}[t!]
    \caption{Delta stage parameters before and after optimization (opt). See corresponding Fig. \ref{fig_delta_before_after_calibration}.}
    \label{tab:optimization_params}
    \begin{center}
        \begin{tabular}{|c|ccccc|}
            \hline
            Parameter & L\,(mm)   & $r_b$\,(mm)   &  $r_p$\,(mm)  &  $\psi$\,(deg) &  $w$\,(mm) \\
            \hline
            Before Opt: &   86 & 54.6 & 24.9 & 21 & 25 \\
            After Opt: & 64 & 61.8 & 32.8 & 27 & 40\\ 
            \hline
        \end{tabular}
    \end{center}
\end{table}

Five different tests were conducted measuring accuracy, repeatability, and deflection. The tests were conducted as follows. First, three workspace tests were conducted where the robot alone (not collaborative control) was tasked to touch points on a grid by via the robot tip in the xz- and yz-planes: an accuracy test with nine points spaced 5\,mm in a 10$\times$10\,mm grid (see Table \ref{tab:error} and Fig. \ref{fig:accuracy_test_10x10}), an accuracy and repeatability test in a 2$\times$2\,mm grid with nine points spaced 1\,mm apart touched five times each (Fig. \ref{fig:accuracy_repeatability_test_2x2}), and a repeatibility and resolution test in a 10$\times$10\,$\mu$m grid with four points spaced at 5\,$\mu$m touched five times each (Fig. \ref{fig:resolution_test_10x10}). For both repeatability and resolution tests (tests two and three), the sequence of motion was programmed such that the movements between the points were random, thus measuring the repeatability of reaching a certain point in workspace from different directions.

The robot tip was then measured in the xy- and yz-plane for repeated roll and tilt joint position tests (Fig. \ref{fig:deflection2}) controlled only by the robot (not collaborative control). The tip was tracked for repeatability tests as individual roll and tilt actuators were commanded to rotate the tip for 0.02$^\circ$.
Finally, a deflection and force versus time test was conducted to show the correlation between various Cartesian forces applied to the robot tool grasping point by a user, and their related displacements. Table \ref{tab:deflection} reports the averaged results from this test with Fig. \ref{fig:deflection1} showing a single instance.

\subsection{Results and Discussion}

\begin{figure}
	\centerline{\includegraphics[width=3.5in]{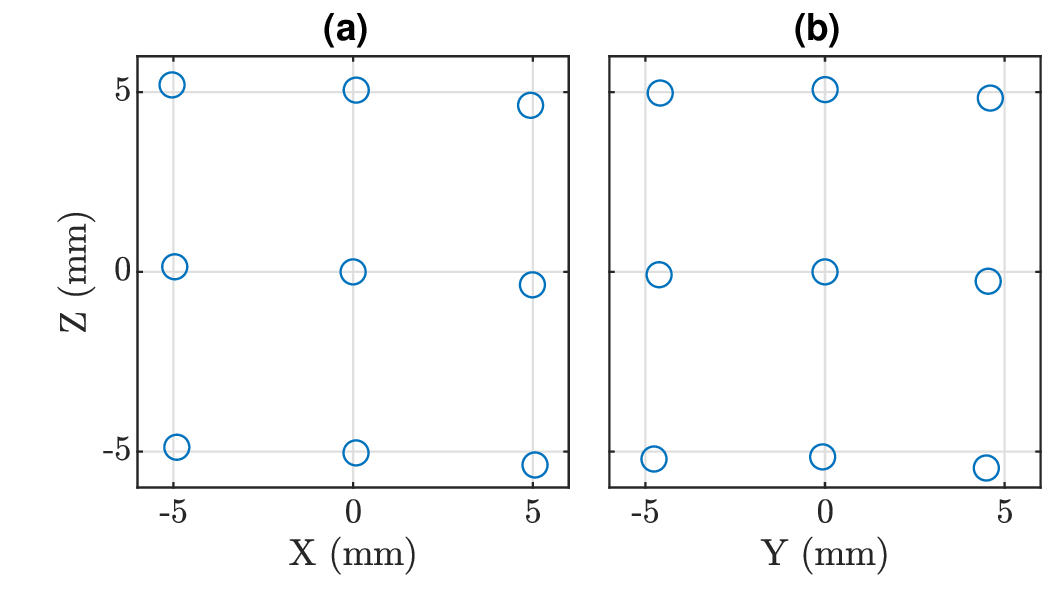}}
	\caption{10$\times$10\,mm workspace relative positioning accuracy test. The robot moved to the nine points at the grid line intersections spaced 5\,mm apart in both the xz- and yz-plane. The circles represent the final achieved position with respect to the intersecting grid lines. Error specifications for this test are reported in Table \ref{tab:error}.}
	\label{fig:accuracy_test_10x10}
\end{figure}

\begin{figure}
	\centerline{\includegraphics[width=3.5in]{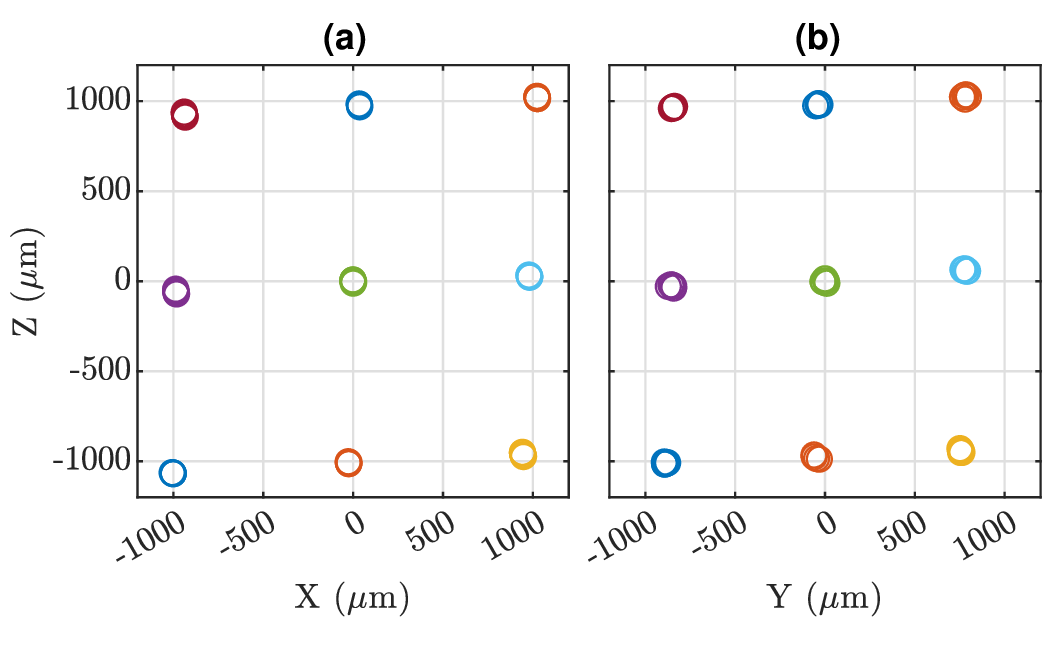}}
	\caption{2$\times$2\,mm workspace accuracy and repeatability test. The robot moved to nine points at the grid line intersections spaced 1\,mm (1000\,$\mu$m) apart in both the xz- and yz-plane repeated five times for each point.}
	\label{fig:accuracy_repeatability_test_2x2}
\end{figure}

\begin{figure}
	\centerline{\includegraphics[width=3.5in]{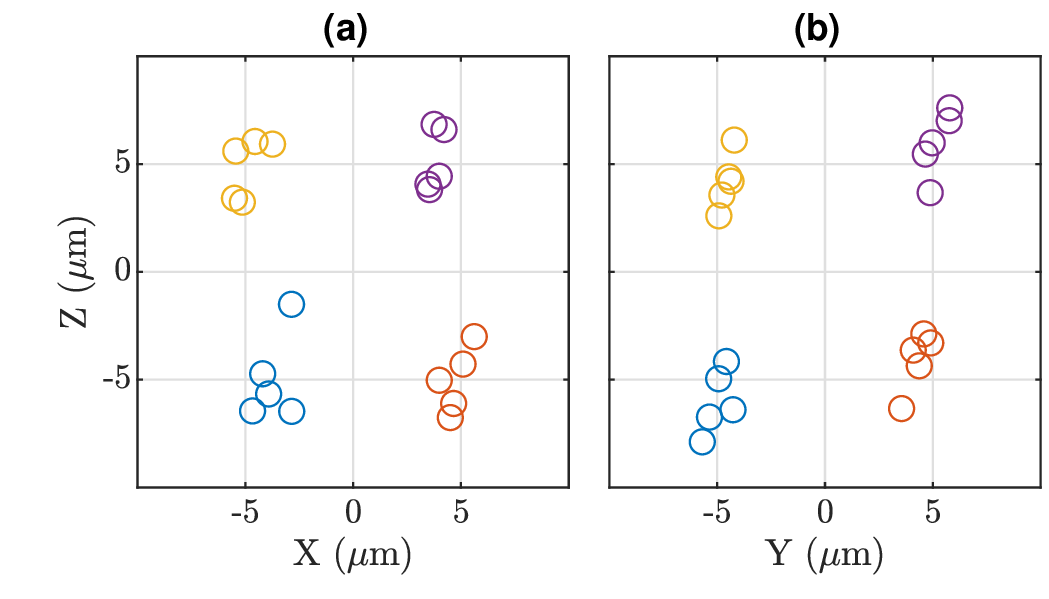}}
	\caption{10$\times$10\,$\mu$m workspace repeatability and resolution test. The robot moved to four points at the grid line intersections spaced 10\,$\mu$m apart in both the xz- and yz-plane repeated five times for each point.}
	\label{fig:resolution_test_10x10}
\end{figure}

\begin{figure}
	\centerline{\includegraphics[width=3.5in]{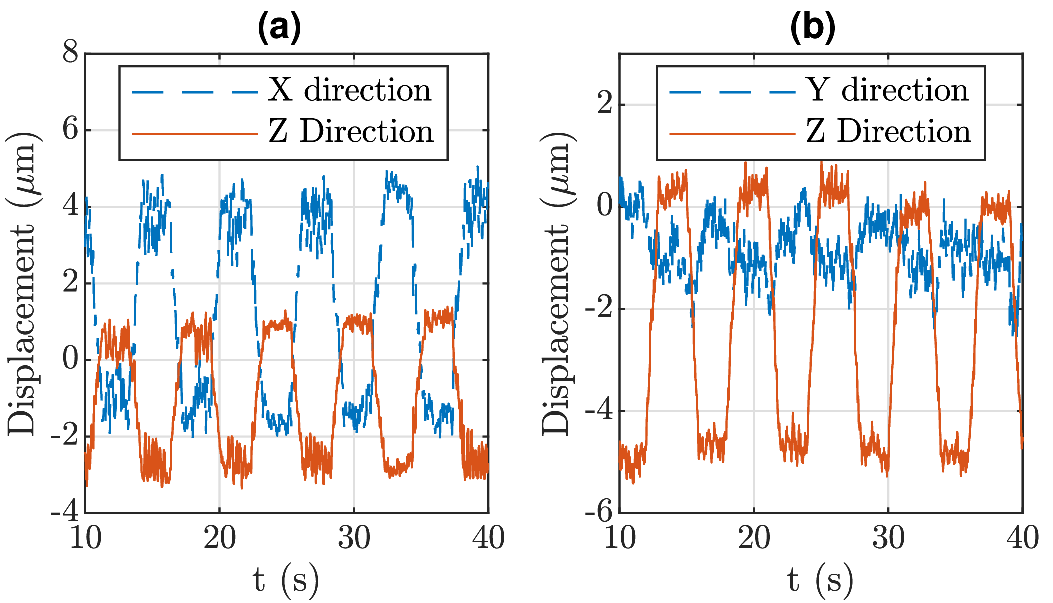}}
	\caption{Rotation and tilt repeatability tests. The roll axis oscillated by 0.02$^\circ$ and positional results for the xz-plane are seen in (a). The tilt axis oscillated by 0.02$^\circ$, and results for the xy-plane are seen in (b).}
        \label{fig:deflection2}
\end{figure}

Results from tests performed are shown in Fig. \ref{fig:accuracy_test_10x10}-\ref{fig:deflection2} and Tables \ref{tab:error}--\ref{tab:deflection}. Table \ref{tab:error} shows the relative positional accuracy errors for motions in the x-, y-, and z-axes of the 10$\times$10\,mm workspace as depicted in Fig. \ref{fig:accuracy_test_10x10}. The first column of the table shows the error in x and z for a motion in the x direction. In other words, for any two points in Fig. \ref{fig:accuracy_test_10x10}(a) that were horizontally aligned, the error between the intended motion of 5\,mm in x and the actual measured motion in x was calculated and averaged over all point pairs. For the x and z error, this lead to a mean and standard deviation of $17\pm2$\,$\mu$m. As only two cameras were used to separately track data in the xz- and yz-planes, entries for xy-plane results do not exist in Table \ref{tab:error}. The highest accuracy for this test was found in motions along the z-axis and the lowest was along the y-axis.

The repeatability for each target point in Fig. \ref{fig:accuracy_repeatability_test_2x2} was defined as the standard deviation of the actual points corresponding to the target point, and then averaged over all nine points. The results show repeatability values of $1.9$\,$\mu$m in x, $9.3$\,$\mu$m in y, and $6.8$\,$\mu$m in z. A conservative estimate from Fig. \ref{fig:resolution_test_10x10} shows that the tip resolution can be defined as 10\,$\mu$m. From a repeatability standpoint, a repeatability of $0.5$\,$\mu$m in x, $0.6$\,$\mu$m in y, and $1.5$\,$\mu$m in z was obtained in this test. It is likely that a smaller resolution is feasible, but achieving this is camera-dependent. 

\begin{figure}
	\centerline{\includegraphics[width=3.5in]{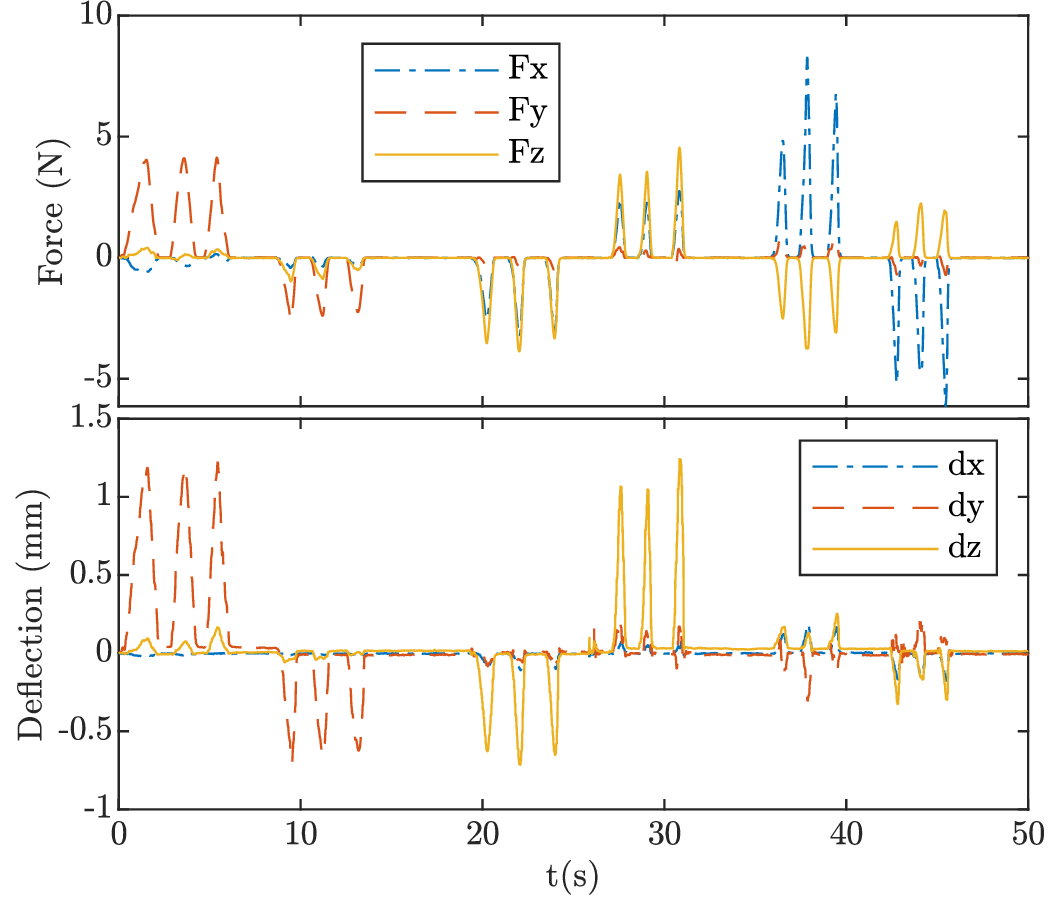}}
	\caption{Deflection and force test versus time. Shown are repeated force deflections in the x-, y-, and z-axes repeated three times each, ranging from approximately -6\,N to 8\,N. Corresponding directly below are the equivalent deflections seen at the robot tool grasping point in the x-, y-, and z-directions as reported by $dx$, $dy$, and $dz$. This figure shows just a single instance of this test. The averages of all tests conducted are reported in Table \ref{tab:deflection}.}
        \label{fig:deflection1}
\end{figure}

The repeatability test in Fig. \ref{fig:deflection2} shows repeatable motions for commanded angular displacements as small as 0.02$^\circ$. This is true for both roll and tilt motions. Deflections in Fig. \ref{fig:deflection1} correlate fairly well with forces applied although scaling between force and displacement is not always proportional between axes. The compliance values from Table \ref{tab:compliance_tests} show the lowest compliance along the x-axis and highest average in the y-axis with the z-axis close behind.

Experimental results also show that the middle of the workspace is more accurate than the outer edges. This is most obvious in Fig. \ref{fig:accuracy_test_10x10}(b) and \ref{fig:accuracy_repeatability_test_2x2}(b) as measurements with a higher y magnitude contain larger errors. Thus correct alignment of the robot initially to the workspace could also improve outcomes. Further modifications could compensate for these errors and errors due to deflection as well.

\begin{table}[b!]
    \caption{10$\times$10\,mm workspace tip accuracy (Fig. \ref{fig:accuracy_test_10x10} data). Error is recorded as mean $\pm$ standard deviation.}
    \label{tab:error}
    \begin{center}
        \begin{tabular}{|c|ccc|}
            \hline
            Error ($\mu m$) & X motion & Y motion & Z motion \\
            \hline
            X & $17\pm2$ & -- & $28\pm7$ \\
            Y & -- & $150\pm10$ & $20\pm7$\\ 
            Z & $4\pm2$ & $31\pm10$ & $12\pm7$\\ 
            \hline
        \end{tabular}
    \end{center}
\end{table}

 Compared to the kinematic specifications defined previously, the SHER 3.0 achieved its accuracy goal for tip accuracy ($<$30\,$\mu$m vs the 25-30\,$\mu$m goal), slightly higher rotational repeatability (0.02$^\circ$ vs 0.01$^\circ$), and slightly higher translational repeatability (conservatively 9.3\,$\mu$m vs 5.0\,$\mu$m). Using more accurate measurement systems, however, could show improvements over these reported values. In comparison to other reported retinal surgery robots reported in \cite{vander2020robotic}, of whom some resolution and repeatability results are reported in Table 2 of \cite{nambi2016compact}, the SHER 3.0 performs similarly. Scores from other robots in this table range from $<$1-10\,$\mu$m for repeatability and 0.2-5\,$\mu$m for resolution.
 
Implementing autonomous tasks requiring high precision (i.e. needle puncture, subretinal injection, etc.) will require utilizing the axis corresponding with the highest accuracy and stiffness. This ensures tasks will be conducted in the safest manner possible. Table \ref{tab:error} shows the lowest error in tip accuracy in the z-direction. Table \ref{tab:deflection} shows the x-axis has the lowest compliance scores. Hence, performing a high-precision task such as needle-puncture would likely be done via the z-axis or x-axis depending on force requirements. For subretinal injection, the direction of puncture requires the highest precision to achieve the proper layer depth and not penetrate iatrogenically further into the retina than needed.

\begin{table}[t]
    \caption{Deflection Table: Shows average compliance at each position $P$ tested.}
    \label{tab:deflection}
    \begin{center}
        \begin{tabular}{|c|ccccc|}
            \hline
            & \multicolumn{5}{c|}{Position of P ($X_p$, $Y_p$)(mm)} \\
            \hline
            Compliance ($\mu m/N$) & (1,1) & (18,12) & (-7,-7) & (-11,3) & (6,-7) \\
            \hline
            $C_x$ & 20 & 23 & 28 & 22 & 23\\ 
            $C_y$ & 270 & 230 & 320 & 290 & 290\\ 
            $C_z$ & 180 & 270 & 200 & 350 & 200 \\
            \hline
        \end{tabular}
    \end{center}
\end{table}

It is important to note that the robot can be used in either an autonomous or cooperative mode. In autonomous mode we do not worry about robot stiffness whereas in cooperative control mode we do.

SHER 3.0 is also less expensive than previous iterations and achieves improved integration into existing operating environments due to a smaller footprint. Its smaller light weight design and mobile passive structure make it particularly appealing for use in confined operating room spaces. From lab use, the authors have observed that it is better coupled to human natural motion, providing improved ergonomics over past SHER platforms.

Along with previous work outlining improved robot ergonomics \cite{wu2020optimized, roth2021towards}, the SHER 3.0 has been used previously in vessel-following analysis tasks \cite{zhao2023human}. Future work planned with the SHER 3.0 aims to test the robot in various scenarios in preparation for applications in retinal surgery. These include tests for multi-user experiments performing tasks for vein cannulation, subretinal injections, membrane peeling and others.  Anticipated \emph{in vivo} animal studies using rabbits and pigs will also help to validate the robot in a live surgical scenario. Additionally, comparisons could be made evaluating whether the current robot outperforms previous SHER robots.




\section{Conclusion}\label{sec:Conclusion}

The SHER 3.0 is a novel 5-DOF robot showcasing robust and accurate parallel and arm structures capable of performing sub-millimeter accuracy tasks sufficient for subretinal injection and other vitreoretinal surgical maneuvers. As an improved iteration on previous collaborative control SHER robots, SHER 3.0 provides a desirable smaller size, optimized torsional compliance and GCI, and offers minimal deflection obtaining sufficient tip accuracies.



\bibliography{Manuscript}

\end{document}